%% file: arxiv_long_version.tex
\documentclass{article}

\usepackage[margin=1.42in]{geometry}

\PassOptionsToPackage{numbers}{natbib}
\usepackage[utf8]{inputenc} 
\usepackage[T1]{fontenc}    
\usepackage{hyperref}       
\usepackage{url}            
\usepackage{booktabs}       
\usepackage{amsfonts}       
\usepackage{nicefrac}       
\usepackage{microtype}      
\usepackage{xfunctions}     
\usepackage{algorithm2e}
\usepackage{subcaption}

\usepackage{color}
\usepackage{xspace}
\usepackage{url}

\include{macros}

\newcommand\blfootnote[1]{%
\begingroup
\renewcommand\thefootnote{}\footnote{#1}%
\addtocounter{footnote}{-1}%
\endgroup
}

\newcommand{\starcraft}{StarCraft\xspace}

\title{Episodic Exploration for Deep Deterministic Policies:\\
An Application to StarCraft Micromanagement Tasks}

\author{
    Nicolas Usunier$^*$, Gabriel Synnaeve$^*$, Zeming Lin, Soumith Chintala\blfootnote{\*: These authors contributed equally to this work.}\\
  Facebook AI Research\\
  \texttt{usunier,gab,zlin,soumith@fb.com}
}

\begin{document}

\maketitle

\begin{abstract}

  We consider scenarios from the real-time strategy game StarCraft as new
  benchmarks for reinforcement learning algorithms. We propose
  \emph{micromanagement} tasks, which present the problem of the short-term,
  low-level control of army members during a battle. From a reinforcement
  learning point of view, these scenarios are challenging because the
  state-action space is very large, and because there is no obvious feature
  representation for the state-action evaluation function. We describe our
  approach to tackle the micromanagement scenarios with deep neural network
  controllers from raw state features given by the game engine. In addition, we
  present a heuristic reinforcement learning algorithm which combines direct
  exploration in the policy space and backpropagation. This algorithm allows for the
  collection of traces for learning using deterministic policies, which appears much
  more efficient than, for example, $\epsilon$-greedy exploration. Experiments show
  that with this algorithm, we successfully learn non-trivial strategies for
  scenarios with armies of up to $15$ agents, where both Q-learning and
  REINFORCE struggle.

\end{abstract}

\section{Introduction}
StarCraft\footnote{StarCraft and its expansion StarCraft: Brood War
  are trademarks of Blizzard Entertainment$^{\mathrm{TM}}$} is a
real-time strategy game in which each player must build an army and
control individual units to destroy the opponent's army. As of
today, StarCraft is considered one of the most difficult games for
computers, and the best bots only reach the level of high amateur
human
players\footnote{\url{http://webdocs.cs.ualberta.ca/\~cdavid/starcraftaicomp/report2015.shtml\#mvm}\\ (retrieved
  on August $23$rd, 2016).}. The main difficulty comes from the need to
control a large number of units in a wide, partially observable
environment. This implies, in particular, extremely large state and
action spaces: in a typical game, there are at least $10^{1685}$
possible states (for reference, the game of Go has about $10^{170}$
states) and the joint action space is in $\Theta((\#\text{commands per
  unit})^{\# \text{units}})$, with a peak number of units of about
$400$ \cite{synnaeve2011bayesian}. From a machine learning point of
view, StarCraft provides an ideal environment to study the control of
multiple agents at large scale, and also an opportunity to define
tasks of increasing difficulty, from \emph{micromanagement}, which
concerns the short-term, low-level control of fighting units during
battles, to long-term strategic and hierarchical planning under
uncertainty. While building a controller for the full game based on
machine learning is out-of-reach for current methods, we propose, as a
first step, to study reinforcement learning algorithms in
micromanagement scenarios in StarCraft.

Both the work on Atari games \cite{mnih2013playing} and the recent Minecraft
scenarios studied by researchers \cite{abel2016exploratory,oh2016control}
focus on the control of a single agent, with a fixed, limited set of actions.
Coherently controlling multiple agents (units) is the main challenge of
reinforcement learning for micromanagement tasks. This comes with two
main difficulties. The first difficulty is to efficiently explore the
large action space.
The implementation of a coherent strategy requires the units to take
actions that depend on each other, but it also implies that any small
alteration of a strategy must be maintained for a sufficiently long
time to properly evaluate the long-term effect of that change. In
contrast to this requirement of consistency in exploration, the
reinforcement learning algorithms that have been successful in
training deep neural network policies such as Q-learning
\cite{watkins1992q,sutton1998reinforcement} and REINFORCE
\cite{williams1992simple,deisenroth2013survey}, perform exploration by
randomizing actions. In the case of micromanagement, randomizing
actions mainly disorganizes the units, which then rapidly lose the
battle without collecting relevant feedback.
The second major difficulty of micromanagement scenarios is that
there is no obvious way to parameterize the policy given the state and
the actions, because some actions describe a relation between entities
of the state, e.g., (unit A, attack, unit B) or (unit A, move,
position B) and are not restricted to a few constant symbols such as
``move left'' or ``move right''. The approach of ``learning directly
from pixels'', in which the pixel input is fed to a multi-class
convolutional neural network, was successful in Atari games
\cite{mnih2015human}. However, pixels only capture spatial
relationships between units. These are only parts of the relationships
of interest, and more generally this kind of multi-class architecture
cannot evaluate actions that are parameterized by an entity of the
state.

The contribution of this paper is twofold. First, we propose several
micromanagement tasks from StarCraft (Section~\ref{sec:starcraft}), then we
describe our approach to tackle them and evaluate well known reinforcement
learning algorithms on these tasks (Section~\ref{sec:framework}), such as
Q-learning and REINFORCE (Subsection~\ref{sec:preliminaries}). In particular,
we present an approach of greedy inference to break out the complexity of
taking the actions at each step (Subsection~\ref{sec:greedy}). We also describe
the features used to jointly represent states and actions, as well as a deep
neural network model for the policy (Section~\ref{sec:model}). Second, we
propose a heuristic reinforcement learning
algorithm to address the difficulty of exploration in these tasks
(Section~\ref{sec:zero_order}). To avoid the pitfalls of exploration by taking
randomized actions at each step, this algorithm explores directly in policy
space, by randomizing a small part of the deep network parameters at the
beginning of an episode and running the altered, deterministic, policy
thoughout the whole episode. Parameter updates are performed using a heuristic
approach combining gradient-free optimization for the randomized parameters,
and plain backpropagation for the others. Compared to algorithms for efficient
direct exploration in parameter space (see e.g.,
\cite{mannor2003cross,sehnke2008policy,szita2006learning,osband2016generalization}),
the novelty of our algorithm is to mix exploration through parameter
randomization and plain gradient descent. Parameter randomization is efficient
for exploration but learns slowly with a large number of parameters, whereas
gradient descent does not take part in any exploration but can rapidly learn
models with millions of parameters.


\section{Related work}

Multi-agent reinforcement learning has been an active area of research (see
e.g., \cite{busoniu2008comprehensive}). Most of the focus has been on learning
agents in competitive environments with adaptive adversaries (e.g.,
\cite{littman1994markov,hu1998multiagent,tesauro2003extending}). Some work has
looked at learning control policies for individual agents in a collaborative
setting with communication constraints
\cite{tan1993multi,bernstein2002complexity}, with applications such as soccer
robot control \cite{stone1999team}, and methods such as hierarchical
reinforcement learning for communicating high-level goals
\cite{ghavamzadeh2006hierarchical}, or learning an efficient communication
protocol \cite{sukhbaatar2016learning}. While the decentralized control
framework is most likely relevant for playing full games of StarCraft, here we
avoid the difficulty of imperfect information, therefore we use the
multi-agent structure only as a means to structure the action space. As in the
approach of \cite{maes2009structured} with reinforcement learning for
structured output prediction, we use a greedy sequential inference scheme at
each time frame: each unit decides on its action based solely on the state
combined with the actions of units that came before it in the sequence.

Algorithms that have been used to train deep neural network controllers in
reinforcement learning include Q-learning \cite{watkins1992q,mnih2015human},
the method of temporal differences
\cite{sutton1988learning,tesauro1995temporal}, policy gradient and their
variants \cite{williams1992simple,deisenroth2013survey}, and actor/critic
architectures
\cite{barto1983neuronlike,silver2014deterministic,silver2016mastering}. Except
for the deterministic policy gradient (DPG) \cite{silver2014deterministic},
these algorithms rely on randomizing the actions at each step for exploration.
DPG collects traces by following deterministic policies that remain constant
throughout an episode, but can only be applied when the action space is
continuous.  Our work is most closely related to works that explore the
parameter space of policies rather than the action space. Several approaches
have been proposed that randomize the parameters of the policy at the beginning
of an episode and run a deterministic policy throughout the entire episode,
borrowing ideas from gradient-free optimization (see e.g.,
\cite{mannor2003cross,sehnke2008policy,szita2006learning}). However, these
algorithms rely on gradient-free optimization for all parameters, which does
not scale well with the number of parameters. Osband et al.
\cite{osband2016generalization} describe another type of algorithm where the
parameters of a deterministic policy are randomized at the beginning of an
episode, and learn a posterior distribution over the parameters as in Thomson
sampling \cite{thompson1933likelihood}. Their algorithm is particularly
suitable for problems in which depth-first search exploration is efficient, so
their motivation is very similar to ours. Their approach was proved to be
efficient, but applies only to linear functions and scales quadratically with
the number of parameters. The bootstrapped deep Q-networks (BDQN)
\cite{osband2016deep} are a practical implementation of the ideas of
\cite{osband2016generalization} for deep neural networks. However, BDQN still
performs exploration in the action space at the beginning of the training, and
there is no randomization of the parameters. Instead, several versions of the
last layer of the deep neural network controller are maintainted, and one of
them is used alternatively during an entire episode to generate diverse traces
and perform Q-learning updates. In contrast, we randomize the parameters of the
last layer once at the beginning of an episode, and contrarily to Q-learning,
our algorithm does not rely on the estimation of the state-action value
function.

In the context of StarCraft micromanagement, a large spectrum of AI approaches
have been studied. There has been work on Bayesian fusion of hand-designed
influence maps \cite{synnaeve2011bayesian}, fast heuristic search (in a
simplified simulator of battles without collisions) \cite{churchill2012fast},
and even evolutionary optimization \cite{liu2014evolving}. Closer to this
work, \cite{wender2012applying} successfully applied tabular Q-learning
\cite{watkins1992q} and SARSA \cite{sutton1998reinforcement}, with and
without experience replay (``eligility traces''), with a reward similar to the
one used in several of our experiments. However, the action space was
reduced to pre-computed ``meta-actions'': fight and retreat, and the features
were hand-crafted. None of these approaches are used \textit{as is} in existing
StarCraft bots, mainly for a lack of robustness to all micromanagement
scenarios that can happen in a full game, for a lack of completeness (both can
be attributed to hand-crafting), or for a lack of computational efficiency
(speed). For a more detailed overview of AI research on StarCraft, the
reader should consult \cite{ontanon2013survey}.

\section{StarCraft micromanagement scenarios}
\label{sec:starcraft}
We focus on micromanagement, which consists of optimizing each unit's
actions during a battle. The tasks presented in this paper represent
only a subset of the complexity of playing StarCraft. As StarCraft is
a real-time strategy (RTS) game, actions are durative (are not fully
executed on the next frame), and there are approximately 24 frames per
second. As we take an action for each unit every few frames
(e.g.~every 9 frames here, see~\ref{sec:specifics} in Appendix for
more details), we only consider actions that can be executed in this
time frame, which are: the 8 move directions, holding the current
position, an attack action for each of the existing enemy units. In
all tasks, we control all units from one side, and the opponent
(built-in AI in the experiments) is attacking us:
\begin{itemize}
    \item \texttt{m5v5} is a task in which we control 5 Marines
      (ranged ground unit), against 5 opponent
      Marines. A good strategy here is to focus fire, by whatever
      means. For example, we can attack the weakest opponent unit (the
      unit with the least remaining life points), with tie breaking, or
      attack the closest to the group.
    \item \texttt{m15v16}: same as above, except we have 15 Marines and the opponent
        has 16. A good strategy here is also to focus fire, while avoiding
        ``overkill'' (spread the damage over several units if the focus firing
        is enough to kill one of the opponent's unit). A Marine has 40 hit
        points, and can hit for 6 hit points every 15 frames.
    \item \texttt{dragoons\_zealots}: symmetric armies with two types of units:
        3 Zealots (melee ground unit) and 2 Dragoons (ranged ground unit). Here a
        strategy requires to focus fire, and if possible to 1) not spend too
        much time having the Zealots walk instead of fight, 2) focus the
        Dragoons (which receive full damage from both Zealots and Dragoons
        while inflicting only half damage on Zealots).
    \item \texttt{w15v17}: we control 15 Wraiths (ranged flying unit)
      while the opponent has 17. Flying units have no ``collision'',
      so multiple units can occupy the same tile.  Here more than
      anywhere, it is important not to ``overkill'': Wraiths have 120
      hit points, and can hit for 20 damage on a 22 frame cooldown. As
      there is no collision, moving is easier.
    \item other \texttt{mXvY} or \texttt{wXvY} scenarios. The 4 scenarios above
        are the ones on which we train our models, but they can learn
        strategies that overfit a given number of units, so we have similar
        scenarios but with different numbers of units (on each side).
\end{itemize}
For all these scenarios, a human expert can win 100\% of the time against the
built-in AI, by moving away units that are hurt (thus conserving firepower) and
with proper focus firing.

\section{Framework: RL and multiple units}
\label{sec:framework}


We now describe the notation and definition underlying the
algorithms Q-learning and policy gradient (PG) used as baselines
here. We then reformulate the joint inference over the potential
actions for different units as a greedy inference which reduces to a
usual MDP with more states but fewer actions per state. We then show
how we normalize cumulative rewards at each state in order to keep
rewards in the full interval $[-1, 1]$, during an entire episode, even
when units disappear.

\subsection{Preliminaries: Q-learning and REINFORCE}
\label{sec:preliminaries}
\paragraph{Notation}
The environment is approximated as an MDP, with a finite set of states
denoted by $\origS$. Each state $\origs$ has a set of units
$\Units{\origs}$, and a policy has to issue a command $\cm\in\Cms$ to
each of them. The set of commands is finite. An action in that MDP is
represented as a sequence of (unit, command) pairs $\act = ((\un{1},
\cm{1}), ..., (\lstun{\origs}, \lstcm{\origs}))$ such that
$\xSet{\un{1}, ..., \lstun{\origs}} =
\Units{\origs}$. $\ssize{\origs}$ denotes the number of units in state
$\origs$ and $\Acts{\origs} =
(\Units{\origs}\times\Cms)^{\ssize{\origs}}$ the set of actions in
state $\origs$. We denote by $\trsit{\origs'|\origs, \act}$ the
transition probability of the MDP and by $\trsit_1$ the probability
distribution of initial states. When there is a transition from state
$\origs^t$ to a state $\origs^{t+1}$, the agent receives the reward $r^{t+1} =
r(\origs^t, \origs^{t+1})$, where $\rew:\origS\times\origS\rightarrow
\Re$ is the reward function. We assume that commands are received and
executed concurrently, so that the order of commands in an action does
not alter the transition probabilities. Finally, we consider the episodic
reinforcement learning scenario, with finite horizon $T$ and
undiscounted rewards. The learner has to learn a (stochastic) policy
$\pol(\act|\origs)$, which defines a probability distribution over
actions in $\Acts{\origs}$ for every $\origs\in\origS$. The objective
is to maximize the expected undiscounted cumulative reward over
episodes $\Erew{\pol} = \expect{\sum_{t=1}^{T-1} r(s^t, s^{t+1})} =
\expect{\cumr{\origs^{1..T}}}$, where the expectation is taken with
respect to $s^1 \sim \trsit_1$, $s^{t+1} \sim \trsit{.|a^t,s^t}$ and
$a^t\sim \pol(.|s^t)$.

We now briefly describe the two algorithms we use as baseline,
Q-learning \cite{sutton1998reinforcement} and REINFORCE
\cite{williams1992simple}.
\paragraph{Q-learning}
The Q-learning algorithm in the finite-horizon setting learns an
action-value function $\qq$ by solving the Bellman equation
\begin{equation}
\label{eq:bellman}
\forall \origs\in\origS, \forall \act\in\Acts{\origs}, \qq_t(\origs, \act)
= \sum_{\origs'\in\origS} \trsit{\origs'|\origs, \act}\big(\rew{\origs, \origs'}
+ \max_{\act'\in\Acts{\origs'}} \qq_{t+1}(\origs', \act')\big)\,,
\end{equation}
where $\qq_t$ is the state-action value function at stage $t$ of an
episode, and $Q_{T}(s,a) = 0$ by convention. $Q_t(s,a)$ is also $0$
whenever a terminal state is reached, and transitions from a terminal
state only go to the same terminal state.

Training is usually carried out by collecting traces $(\origs^t,
\act^t, \origs^{t+1}, r^{t+1})_{t=1, ..., T-1}$ using
$\epsilon$-greedy exploration: at state $\origs$ and stage $t$, an
action in $\argmax_{\act\in\Acts{\origs}} \qq_t(\origs, \act)$ is
chosen with probability $1-\epsilon$, or an action in $\Acts{\origs}$
is chosen uniformly at random with probability $\epsilon$. In
practice, we use stationary $\qq$ functions (i.e., $\qq_t =
\qq_{t+1}$), which are neural networks, as described in Section
\ref{sec:model}. Training is carried out using the standard online
update rule for Q learning with function approximation (see e.g.,
\cite{mnih2015human}), which we apply in mini-batches (see Section
\ref{sec:hyperparams} for more details).

This training phase is distinct from the test phase, in which we
record the average cumulative reward of the deterministic
policy
$\origs \mapsto
\argmax_{\act\in\Acts{\origs}} \qq{\origs, \act}$.

\paragraph{REINFORCE}
The algorithm REINFORCE belongs to the family of policy gradient algorithms
\cite{sutton1999policy}. Given a stochastic policy $\parpol$
parameterized by $\parpolpars$, learning is carried out by generating
traces $(\origs^t, \act^t, \origs^{t+1}, r^{t+1})_{t=1, ..., T-1}$ by
following the current policy. Then, stochastic gradient updates are
performed, using the gradient estimate:
\begin{equation}
\label{eq:reinforce}
\sum_{t=1}^T \cumr{\origs^{t..T}}\nabla_{\parpolpars} \log(\parpol{\act^t|\origs^t})\,.
\end{equation}
We use a Gibbs policy (with temperature parameter $\tau$) as the stochastic
policy:
\begin{equation}
\parpol{\act|\origs} = \frac{\exp(\phi_\Theta(\act,\origs) / \tau)}{\sum_{b \in
\Acts{\origs}}\exp(\phi_\Theta(b,\origs) / \tau)}\,,
\end{equation}
where $\phi_\Theta$ is a neural network with paramters $\Theta$ that
gives a real-valued score to each (state, action) pair. For testing, we
use the deterministic policy $\parpol{\origs} =
\argmax_{\act\in\Acts{\origs}} \phi_\Theta(a,\origs)$.

\subsection{The MDP for greedy inference}
\label{sec:greedy}
One way to break out the complexity of jointly infering the commands
to each individual unit is to perform greedy inference at each step:
at each state, units choose a command one by one, knowing the commands
that were previously taken by other units. Learning a greedy policy
boils down to learning a policy in another MDP with fewer
actions per state but exponentially more states, where the additional
states correspond to the intermediate steps of the greedy
inference. This reduction was previously proposed in the context of
structured prediction by Maes et al. \cite{maes2009structured}, who
proved that an optimal policy in this new MDP has the same cumulative
reward as an optimal policy in the original MDP.

A natural way to define the MDP associated with greedy inference,
hereafter called greedy MDP, is to define the set of atomic actions of
the greedy policy as all possible (unit, command) pairs for the units
whose command is still not decided.  This would lead to
an inference with quadratic complexity with respect to the number of
units, which is undesirable.

Another possibility is to first choose a unit, then a command to apply
to that unit, which yields an algorithm with $2|\origs|$ steps for
state $\origs$. Since the commands are executed concurrently by the
environment after all commands have been decided, the cumulative
reward does not depend on the order in which we choose the
units. Going further, we can let the environment in the greedy MDP
choose the next unit, for instance, uniformly at random among
remaining units. The resulting inference has a complexity that is
linear in the number of units.
More formally, using the notation
$\act_{1..k}$ to denote the $k$ first (unit, command) pairs of an
action $\act$ (with the convention $\act_{1..0} = \emptyset$), the
state space $\greeS$ of the greedy MDP is defined by
\begin{equation*}
\greeS = \xSet[\big]{(\origs, \act_{1..k}, \un{k+1}) \given s \in
  \origS, 0\leq k < \ssize{\origs}, \act =
  ((\un{1}, \cm{1}), ..., (\lstun{\origs}, \lstcm{\origs})) \in
  \Acts{\origs} }.
\end{equation*}
The action space $\gActs{\grees}$ of each state $\grees\in\greeS$ is
constant and equal to the set of commands $\Cms$. Moreover, for each
state $\origs$ of the original MDP, any action $\act = ((\un{1},
\cm{1}), ..., (\lstun{\origs}, \lstcm{\origs}) \in \Acts{\origs}$, the
transition probabilities $\gtrsit$ in the greedy MDP are defined by
\begin{align}
 \forall k\in\xSet{0, ..., \ssize{\origs}-1}, &~~\gtrsit{(\origs,
  \act_{1..k}, \un{k+1}) \big |(\origs, \act_{1..k-1}, \un{k}),
  \cm{k}} = \frac{1}{\ssize{s} - k}\\
\text{~~and~~} \forall \origs'\in\origS, \forall \un'\in\Units{\origs'}, & ~~
\gtrsit{(\origs', \emptyset, \un') \big| (\origs, \act_{1..\ssize{s}-1}, \lstun{\origs}),
  \lstcm{\origs}} = \frac{1}{\ssize{\origs'}} \trsit{\origs'| \origs, \act}\,.
\end{align}
Finally, using the same notation as above, the reward function $\grew$ between states that represent
intermediate steps of the algorithm is $0$ and the last unit to play receives the reward:
\begin{equation*}
\grew{(\origs, \act_{1..k-1}, \un{k}), (\origs,
  \act_{1..k}, \un{k+1})} = 0\,,
~~\text{and}~~ \grew{(\origs, \act_{1..\ssize{s}-1}, \lstun{\origs}), (\origs', \emptyset, \un')}
= \rew{\origs, \origs'}\,.
\end{equation*}
It can be shown that an optimal policy for this greedy MDP chooses
actions that are optimal for the original MDP, because the immediate
reward in the original MDP does not depend on the order in which the
actions are taken. This result only applies if the family of policies
has enough capacity. In practice, some ordering may be easier to learn
than others, but we did not investigate this issue because the gain,
in terms of computation time, of the random ordering was critical for
the experiments.

\subsection{Normalized cumulative rewards}
\label{sec:normRew}
Immediate rewards are necessary to provide feedback that guides
exploration.  In the case of micromanagement, a natural reward signal
is the difference between damage inflicted and incurred between two
states. The cumulative reward over an episode is the total damage
inflicted minus the total damage incurred along the episode. However,
the scale of this quantity heavily depends on the number of units
(both our units and enemy units) that are present in the state, a
quantity which significantly decreases along an episode. Without
proper normalization with respect to the number of units in the
current state, learning will be artificially biased towards the large
immediate rewards at the beginning of the episode.

We present a simple method to normalize the immediate rewards on a
per-state basis, assuming that a scale factor $\sca{\origs}$ is
available to the learner -- it can be as simple as the number of units.
%
%
Then, instead of considering cumulative rewards from a starting state
$\origs^t$,
we define normalized cumulative rewards $\nr^{t..T}$ as the following
recursive computation over an episode:
\begin{equation*}
\forall t \in \xSet{1, ..., T-1},
\nr^{t..T} = \frac{\rew^{t+1}+\sca{\origs^{t+1}} \nr^{t+1..T}}{\sca{\origs^t}}\,.
\end{equation*}
These normalized rewards maitain the invariant $\nr^{t..T} =
\frac{{\cumr}^{t..T}}{\sca{\origs^t}}$; but more importantly, the
normalization can be applied to the Bellman equation
\eqref{eq:bellman}, which becomes
\begin{equation*}
\forall \origs\in\origS, \forall \act\in\Acts{\origs}, \qq{\origs, \act}
= \sum_{\origs'\in\origS} \frac{\trsit{\origs'|\origs, \act}}{\sca{\origs}}\big(\rew{\origs, \origs'}
+ \sca{\origs'}\max_{\act'\in\Acts{\origs'}} \qq{\origs', \act'}\big)\,.
\end{equation*}

The stochastic gradient updates for Q-learning can easily be modified
accordingly, as well as the gradient estimate in REINFORCE
\eqref{eq:reinforce} in which we replace $\cumr$ by $\nr$.

One way to look at this normalization process is to consider that the
reward is $\frac{\rew^{t+1}}{\sca{\origs^t}}$, and
$\frac{\sca{\origs^{t+1}}}{\sca{\origs^t}}$ plays the role of an
(adaptive) discount factor, which is chosen to be at most $1$, and
strictly smaller than 1 when the number of units change.

\section{Features and model for micromanagement in \starcraft}
\label{sec:model}


The features and models we use are intended to test the ability of RL
algorithms to learn strategies when given as little
prior knowledge as possible. We voluntarily restrict ourselves to raw
features extracted from the state description given by the game
engine, without encoding any prior knwoledge of the game
dynamics. This contrast with prior work on Q-learning for
micromanagement such as \cite{wender2012applying}, which use features
such as the expected inflicted damage. We do not allow ourselves
to encode the effect of an attack action on the hit points
of the attacked unit; we do not, either, construct cross-features nor
provide any relevant discretization of the features (e.g., whether
unit A is in the range of unit B). The only transformation of the raw
features we perform is the computation of distances between units and
(between) targets of commands.

We represent a state as a sequence of feature vectors, one feature
vector per unit (ally or enemy) in the state. We remind that each
state in the greedy MDP is a tuple $\grees = (\origs, \act_{1..k},
\un{k+1})$ and an action in that MDP corresponds to a command $\cm$
that $\un{k+1}$ shall execute. At each frame and for each unit, the
commands we consider are (1) attack a given enemy unit, and (2) move
to a specific position. In order to reduce the number of possible move
commands, we only consider $9$ move commands, which either correspond
to a move in one of the 8 basic directions, or staying at the same
position.

Attack commands are non-trivial to featurize because the model needs
be able to solve the reference from the identifiers of the units that
attack or are attacked to their corresponding attributes. In order to
solve this issue, we construct a joint state/action feature
representation in which the unit positions (coordinates on the map)
are indirectly used to refer to the units. We now detail the feature
representation we use and the neural network model.

\subsection{Raw state information and featurization}

For each unit (ally or enemy), the following attributes are extracted
from the raw state description given by the game engine:
\begin{enumerate}
\item Unit attributes: the unit \emph{type}, its coordinates on the
  map (\emph{pos}), the remaining number of hit points (\emph{hp}),
  the \emph{shield}, which corresponds to additional hit points that
  can be recovered when the unit is not attacked, and, finally the
  weapon cooldown (\emph{cd}, number of frames to wait to be able to
  inflict damage again). An additional flag \emph{enemy} is
  used to distinguish between our units and enemy units.
\item Two attributes that describe the command that is currently
  executed by the unit. First, the \emph{target} attribute, which, if
  not empty, is the identifier of the enemy unit currently under
  attack. This identifier is an integer that is attributed arbitrarily
  by the game engine, and does not convey any semantics. We do not
  encode directly this identifier in the model, but rather only use
  the position of the (target) unit (as we describe below in the
  distance features).

  The second attribute, \emph{target\_pos} gives the
  coordinates on the map of the position of the target (the desired
  destination if the unit is currently moving, or the position of the
  \emph{target} if the latter is not empty). These fields are
  available for both ally and enemy units; from these we infer the
  current command \emph{cur\_cmd} that the unit currently performs.
\end{enumerate}

In order to assign a score to a tuple $((\origs, \act_{1..k},
\un{k+1}), \cm)$ where $\cm$ is a candidate command for $\un{k+1}$,
the joint representation is defined as sequence of feature vectors,
one for each unit $\un\in\Units{\origs}$. The feature vector for unit
$\un$, which is denoted by $F(\un, \act_{1..k}, \un{k+1}, \cm)$, is
a joint representation of $\un$ together with its next command \emph{next\_cmd} if it
has already been decided (i.e. if $\un$ is an ally unit whose next
command is in $\act_{1..k}$), and of the command $\cm$ that is
evaluated for $\un{k+1}$. All commands have a field \emph{act\_type}
(attack or move) and a field \emph{target\_pos}. If we want to
featurize a command that is not available for a given unit, such as
the next command for an enemy unit, we set \emph{act\_type} to a
specific ``no command'' value and \emph{target\_pos} to the unit
position.

Given $\un\in\Units{\origs}$, the vector $F(\un, \act_{1..k}, \un{k+1}, \cm)$
contains the $17$ features described below. We use an object-oriented
programming notation ``a.b'' to refer to the value of attribute b of a:
\paragraph{Non-positional features} $\un.enemy$ (boolean), $\un.type$
(categorical, one-hot encoding), $\un.hp$, $\un.shield$, $\un.cd$ (all
three real-valued), $\un.cur\_cmd.act\_type$ (categorical, one-hot
encoding), $\un.next\_cmd.act\_type$, $(\un{k+1}).type$. At this
stage, we do not encode the type of the command
$\cm.$\emph{act\_type}, which is another input to the network (see
Section \ref{sec:deep}).
\paragraph{Relative distance features} $\norm{a - b}$ for \\
$a \in \xSet{\un.pos, \un.cur\_cmd.target\_pos,
  \un.next\_cmd.target\_pos}$ and \\ $b\in\xSet{\un{k+1}.pos,
  \un{k+1}.cur\_cmd.target\_pos, \cm.target\_pos}$.  These features,
in particular, encode which unit is $\un{k+1}$ because the distance
between positions is $0$. They also encode which unit is the target of
the command, and which units have the same target. This encoding is
unambiguous as long as units cannot have the same position, which is
not true for flying units (units have the same position, either as
actor or target of a command, will be treated as the same). In
practice however, the confusion of units did not seem to be a major
issue since units rarely have exactly the same position.

Finally, the full (state, action) tuple of the greedy MDP $((\origs,
\act_{1..k}, \un{k+1}), \cm)$ is represented by an
$\ssize{\Units{\origs}} \times 17$ matrix, in which the $j$-th row is
$F(\un{j}, \act_{1..k}, \un{k+1}, \cm)$. The model, which we describe
below, deals with the variable-size input with global pooling
operations.

\subsection{Deep Neural Network model}
\label{sec:deep}
As we shall see in Section \ref{sec:zero_order}, we consider
state-action scoring functions of the form\\ $\argmax_{c\in\Cms}
\xdp{w, \Psi_\theta((\origs, \act_{1..k}, \un{k+1}),\cm)}$, where $w$
is a vector in $\Re^d$ and $\Psi_\theta((\origs, \act_{1..k},
\un{k+1})\cm)$ is an deep network with parameters $\theta$ which
embeds the state and command of the greedy MDP into $\Re^d$.

The embedding network takes as input a $\ssize{\Units{\origs}} \times
17$ matrix, which we describe below, and operates in two steps:

{\bf (1) Cross featurization and pooling} in this step, each row
$F(\un{j}, \act_{1..k}, \un{k+1}, \cm)$ goes through a 2-layer neural
network, with each layer of width $100$, with an ELU nonlinearity
\cite{clevert2015fast} for the first layer and hyperbolic tangeants as
final activation functions. The resulting $\ssize{\Units{\origs}}
\times 100$ matrix is then aggreated into two different vectors of
size $100$: the first one by taking the mean value of each column
(average pooling), and the second one by taking the maximum (max
pooling). The two vectors are then concatenated and yield a
$200$-dimensional vector for the next step. We can note that this
final fixed-length representation is invariant to the ordering of rows
of the original matrix.

{\bf (2) Scoring with respect to action type} the $200$-dimensional
vector is then concatenated with the type of action $\cm.act\_type$
(one-hot encoding of two values: attack or move). The concatenation
goes through a 2-layer network with $100$ activation units at each
layer. The first non-linearity is an ELU, while the second is a
rectifier linear unit (ReLU).

The rationale behind this model is that it can represent the answer
to a variety of question regarding the relationship between the
candidate command and the state, such as: what is the type of unit of
the command's target? How many damages shall be inflicted? How many
units already have the same target? How many units are attacking
$\un{k+1}$?

Yet, in order to answer these questions, the learner must perform the
appropriate cross-features and paramter updtaes from the reinforcement
signal alone, so the learning task is non-trivial even for fairly
simple strategies.

\section{Combining backpropagation and a zero-order gradient estimates}
\label{sec:zero_order}

We now present our algorithm for exploring deterministic policies in
discrete action spaces, based on policies parameterized by a deep
neural network. Our algorithm is inspired by finite-difference methods
for stochastic gradient-free optimization
\cite{kiefer1952stochastic,nemirovsky1982problem,spall1997one} as well
as exploration strategies in parameter space
\cite{ruckstiess2010exploring}. This algorithm can be viewed as a
heuristic. We present it within the general MDP formulation of Section
\ref{sec:preliminaries} for simplicity, although our experiments apply
it to the greedy MDP of Section \ref{sec:greedy}.


As described in Section \ref{sec:deep}, we consider the case where
pairs (state, action) $(\origs, \act)$ are embedded by a parametric
function $\Psi_\theta(\origs, \act)\in \Re^d$. The deterministic
policy is parameterized by an additional vector $w\in\Re^d$, so that
the action $\pi_{w, \theta}(\origs)$ taken at state $\origs$ is
defined as
\begin{equation*}
\pi_{w, \theta}(\origs) = \argmax_{\act\in\Acts{\origs}} \xdp{w, \Psi_\theta(\origs, \act)}\,.
\end{equation*}

The overall algorithm is described in Algorithm \ref{alg}. In order to
explore the policy space in a consistent manner during an episode, we
uniformly sample a vector $u$ on the unit sphere and run the policy
$\pi_{w+\delta u, \theta}$ for the whole episode, where $\delta > 0$ is
a hyper-parameter.

In addition to implementing a local random search in the policy space,
the motivation for this randomization comes from stochastic
gradient-free optimization
\cite{kiefer1952stochastic,nemirovsky1982problem,spall1997one,duchi2013optimal,ghadimi2013stochastic},
where the gradient of a differentiable function $x\in\Re^d\mapsto
f(x)$ can be estimated with finite difference methods by
\begin{equation*}
\nabla f(x) \approx \expect{\frac{d}{\delta}f(x+\delta u) u}\,,
\end{equation*}
where the expectation is taken over the vector $u$ sampled on the unit
sphere \cite[chapter 9.3]{nemirovsky1982problem}. The constant
$\frac{d}{\delta}$ will be absorbed by learning rates, so we ignore it
in the following. Thus, given a (state, action) pair $(\origs, \act)$
and the observed cumulative reward $\cumr$, we use $\cumr u$ as an
estimator of the gradient of the expected cumulative reward with
respect to $w$ (line (*) \ref{alg}).

The motivation for the update of the network parameters is the
following: given a function $(w\in\Re^d, v\in\Re^d) \mapsto g(\xdp{w,
  v})\in\Re$, we have $\nabla_w g = g'(\xdp{w, v}) v$ and $\nabla_v g
= g'(\xdp{w, v}) w$. Denoting by $\frac{w}{v}$ the term-by-term
division of vectors (assuming $v$ contains only non-zero values) and
$\odot$ the term-by-term multiplication operator, we obtain $\nabla_v
g = (\nabla_w g) \odot \frac{w}{v}$. The update (**) in the algorithm
corresponds to taking $v=\Psi_\theta(\origs, \act)$ in the above, and
using $\cumr u$ as the estimated gradient of the cumulative reward
with respect to $w$, as before.  Since we need to make sure that the
ratios $\frac{w}{\Psi_\theta(\origs, \act)}$ are bounded, in practice
we use the sign of $\frac{w}{\Psi_\theta(\origs, \act)}$ to avoid
numerical issues. This ``estimated'' gradient is then backpropagated
through the network. Preliminary experiments suggested that taking the
sign was as effective as e.g., clipping, and was simpler since there
is no parameter, so we use this heuristic in all our experiments.

The reasoning above is only a partial justification of the update
rule (**) of Algorithm \ref{alg}, because we neglected the dependency
between the parameters and the $\argmax$ operation that chooses the
actions. Nonetheless, considering (**) as a crude approximation to
some real estimator of the gradient seems to work very well in
practice, as we shall see in our experiments. Finally, we use Adagrad
\cite{duchi2011adaptive} to update the parameters of the different
layers. We found the use of Adagrad's update scheme fairly important
in practice, compared to other approaches such as
RMSProp \cite{Tieleman2012}, even though RMSProp tended to work
slightly better with Q-learning or REINFORCE in our experiments.

{\small
\begin{algorithm}[t]
\SetAlgoLined
{\bf Input} exploration hyper-parameter $\delta$, learning rate $\eta$,\\
(state, action) embedding network $\Psi_\theta(\origs, \act)$
taking values in $\Re^d$, with parameters $\theta\in\Re^m$.\;
{\bf initialization}: $w \leftarrow 0$, $w\in\Re^d$\;
\While{stopping criterion not met}{
  Sample $u$ uniformly on the unit sphere of $\Re^d$\;

  $t\leftarrow 1$ \tcp{follow the perturbed deterministic policy for one episode}
  \While{episode not ended}{
    $t \leftarrow t+1$\;
    observe current state $\origs^t$ and reward $r^t$\;
    choose action $\act^t = \argmax_{\act\in\Acts{\origs}} \xdp{w+\delta.u,\Psi_\theta(\origs^t, \act)}$\;
  }
  $\hat{g}(w) = 0$ \tcp{estimate of gradient of the cumul. reward w.r.t. $w$}
  $\hat{G}(\theta) = 0\in\Re^{m\times d}$ \tcp{estimate of gradient of the cumul.
    reward w.r.t. $\theta$}
  $R = 0$ \tcp{cumulative reward}
  \For{$k=t-1$ \KwTo $1$}{
    $R = R + r^{k+1}$ \tcp{use update of Section \ref{sec:normRew} for normalized rewards}
    $\hat{g}(w) \leftarrow \hat{g}(w) + \frac{R}{t}u$\;
    $\hat{G}(\theta) \leftarrow \hat{G}(\theta) + \frac{R}{t}u \odot \big(\sign\frac{w}{\Psi_\theta(\origs^k, \act^k)}\big)$\;
  }
  \tcp{perform gradient ascent}
  update\_adagrad($w$, $\eta\hat{g}(w)$)  \quad\quad\quad(*)\;
  update\_adagrad($\theta$, $\eta\hat{G}(\theta)$)\quad\quad~~(**)\;
}
\caption{Zero-order (ZO) backpropagation algorithm}
\label{alg}
\end{algorithm}
}

\section{Experiments}
\label{sec:experiments}
\subsection{Setup}

We use Torch7\footnote{\url{www.torch.ch}} for all our experiments. We connect
our Torch code and models to StarCraft through a socket server. 
We ran experiments with deep Q networks (DQN)
\cite{mnih2013playing}, policy gradient (PG) \cite{williams1992simple}, and
zero order (ZO). 
We did an extensive hyper-parameters search, in
particular over $\epsilon$ (for epsilon-greedy exploration in DQN), $\tau$ (for
policy gradient's softmax), learning rates, optimization methods, RL algorithms
variants, and potential annealings. See \ref{sec:hyperparams} in Appendix for
more details.

\subsection{Baseline heuristics}
As all the results that we report are against the built-in AI, we compare our
win rates to the ones of (strong) baseline heuristics. Some of these heuristics
often perform the micromanagement in full-fledged StarCraft bots
\cite{ontanon2013survey}, and are the basis of heuristic search
\cite{churchill2012fast}. The baselines are the following:
\begin{itemize}
    \item \textit{random no change} (rand\_nc): select a random target for each
        of our units and do not change this target before it dies (or our unit
        dies). This spreads damage over several enemy units, and can be rather
        bad when there are collisions (because it can require our units to move
        a lot to be in range of their target).
    \item \textit{noop}: literally send no action, that is something that is
        forbidden for our models to do. In this case, the built-in AI
        will control our units, so this exhibit the symmetry (or not!) of a
        given scenario. As we are always in a defensive position, with the
        enemy commanded to walk towards us, all other things considered equal (number of
        units), it should be easier for the defending built-in AI than for the
        attacking one.
    \item \textit{closest} (c): each of our units targets the enemy
      unit closest to it. This is not a bad heuristic as enemy units formation
      (because of collisions) will always make it so that several of
      our units have the same opponent unit as closest unit (some form
      of focus firing), but not all of them (no overkill). It is also
      quite robust for melee units (e.g. Zealots) as it means they
      spend less time moving and more time attacking.
    \item \textit{weakest closest} (wc): each of our units targets the weakest
        enemy unit. The distance of the enemy unit to the center of mass of our
        units is used for tie-breaking. This may overkill.
    \item \textit{no overkill no change} (nok\_nc): same as the weakest closest
        heuristic, but register the number of our units that target each
        opponent unit, choosing another target to focus fire when it becomes
        overkill to keep targeting a given unit. Each of our units keep firing
        on their target without changing (that would lead to erratic behavior).
        Note that the ``no overkill'' component of the heuristic cannot easily
        take the dynamics of the game into account, and so if our units die
        without doing their expected damage on their target, ``no overkill''
        can be detrimental (as it is implemented).
\end{itemize}

\subsection{Results}

\begin{figure}
    \caption{Example of the training uncertainty (one standard deviation) on 5
    different initialization for DQN (left) and zero-order (right) on the m5v5
    scenario.}
    \centering
    \includegraphics[width=0.48\columnwidth]{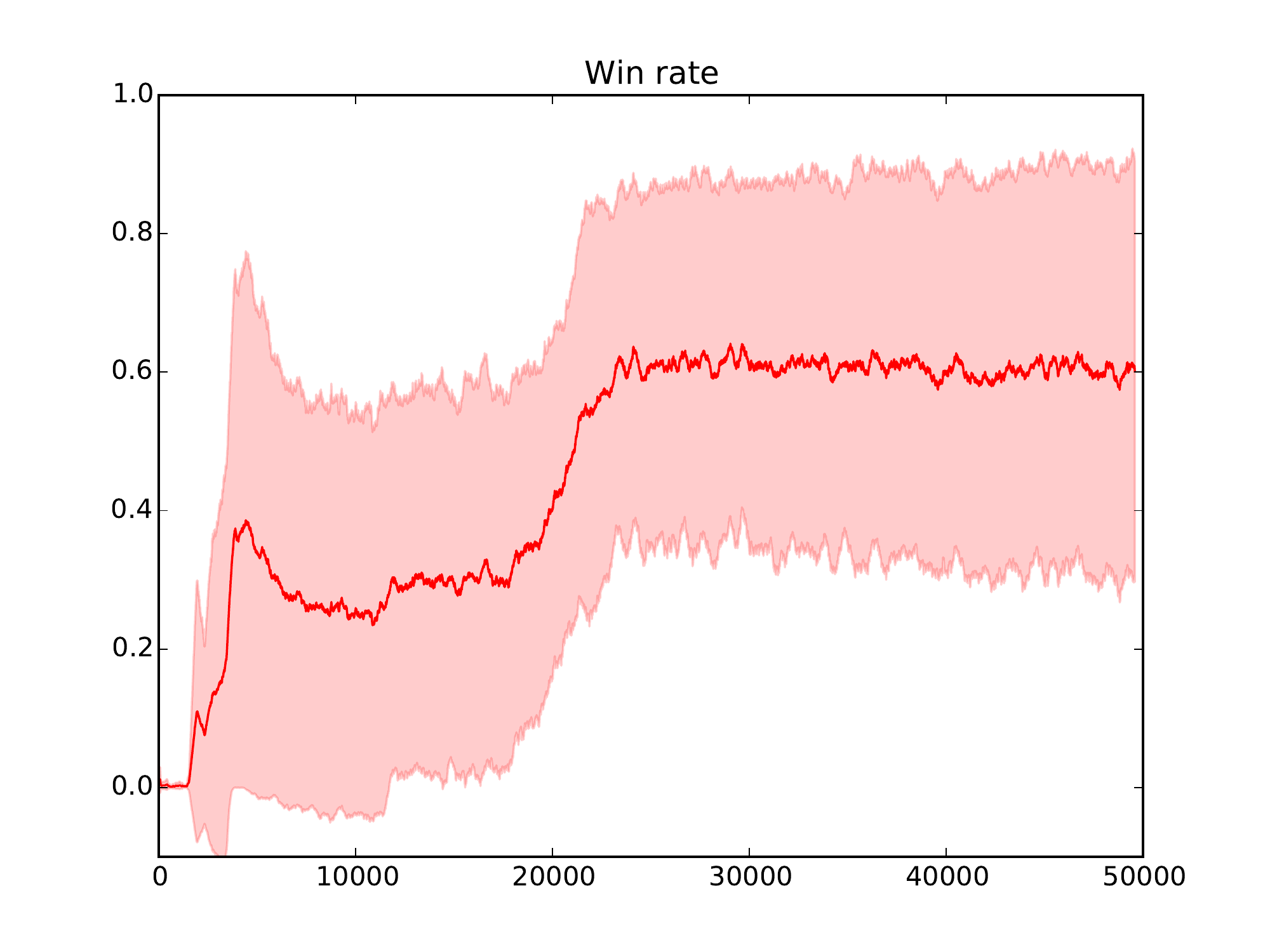}
    \quad
    \includegraphics[width=0.48\columnwidth]{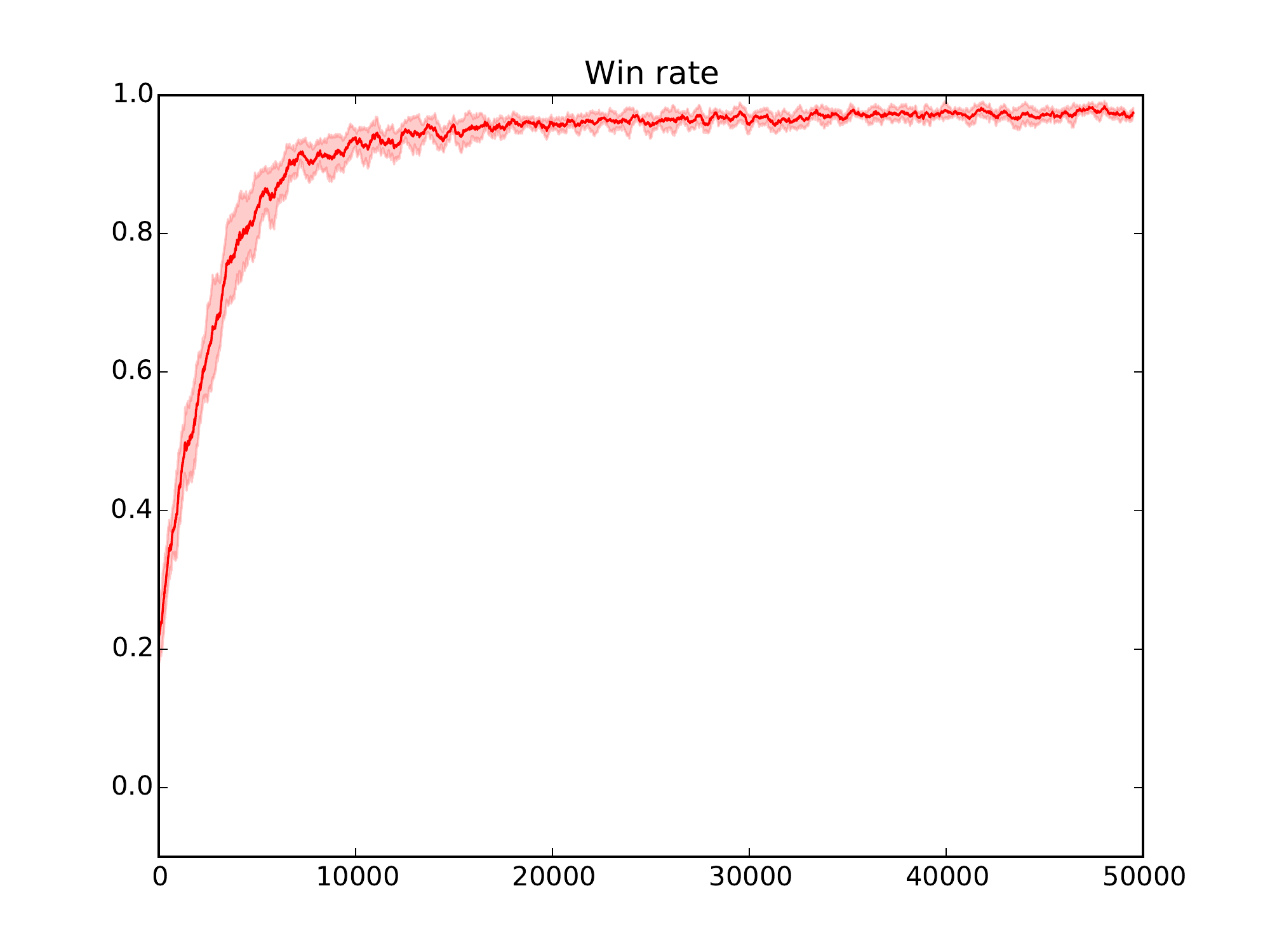}
    \label{fig:training_uncertain}
\end{figure}

The first thing that we looked at were sliding average win rates (over 400
battles) during training against the built-in AI of the various models. In
Figure~\ref{fig:training_uncertain}, we can see than DQN is much more dependent
on initialization and variable (fickling) than zero order (ZO). DQN can
unlearn, reach suboptimal plateaux, or overall need a lot of exploration to
start learning (high sample complexity).


For all the results that we present in Tables \ref{tbl:test_in_domain}
and \ref{tbl:test_out_of_domain}, we ran the models in ``test mode''
by making them deterministic. For DQN we remove the epsilon-greedy
exploration (set $\epsilon=0$), for PG we do not sample in the Gibbs
policy but instead take the value-maximizing action, and for ZO we do
not add noise to the last layer.

We can see in Table~\ref{tbl:test_in_domain} that m15v16 is at the
advantage of our player's side (\textit{noop} is at 81\% win rate), whereas
w15v17 is hard (\textit{c} is at 20\% win rate). By looking
just at the results of the heuristics, we can see that overkill is a
problem on m15v16 and w15v17 (\textit{nok\_nc} is better than
\textit{wc}). ``Attack closest'' (\textit{c}) is approximatively as
good as \textit{nok\_nc} at spreading damage, and thus better on
m15v16 because there are lots of collisions (and attacking the closest
unit is going to trigger less movements).

Overall, the zero order optimization outperforms both DQN and PG
(REINFORCE) on most of the maps. The only map on which DQN and PG
perform well is m5v5. It seems to be easier to learn a focus firing
heuristic (e.g. ``attack weakest'') by identifying and locking on a
feature, than to also learn not to ``overkill''.

\setlength{\tabcolsep}{4pt}
\begin{table}[t]
    \caption{Test win rates over 1000 battles for the training scenarios, for all methods and for heuristics baselines. The best result for a given map is in bold.}
    \vspace*{-0.55cm}
    \label{tbl:test_in_domain}
    \begin{center}
    \begin{tabular}{l ccccc ccc}
        \toprule
         & \multicolumn{5}{c}{heuristics} & \multicolumn{3}{c}{RL} \\
         \cmidrule(lr){2-6} \cmidrule(lr){7-9}
        map & rand\_nc & noop & c & wc & nok\_nc & DQN & PG & ZO \\
        \midrule
        dragoons\_zealots               & .14      & .49          & .67         & .83 & .50     & .61           & .69 & \textbf{.90} \\
        m5v5                            & .49      & .84          & .94         & .96 & .83     &  \textbf{.99 }& .92 & \textbf{1.} \\
        m15v16                          & .00      & \textbf{.81} & \textbf{.81}& .10 & .68     & .13           & .19 & \textbf{.79} \\
        w15v17                          & .19      & .10          & .20         & .02 & .12     & .16           & .14 & \textbf{.49} \\
        \bottomrule

    \end{tabular}
    \end{center}
\end{table}

We then studied how well a model trained on one of the previous maps
performs on maps with a different number of units, to test
generalization.  Table~\ref{tbl:test_out_of_domain} contains the
results for this experiment. We observe that DQN performs the best on
m5v5 when trained on m15v16, because it learned a simpler (but more
efficient on m5v5) heuristic. ``Noop'' and ``attack closest'' are
quite good with the large Marines map because they generate less moves
(and less collisions). Overall, ZO is consistently significantly
better than other RL algorithms on these generalization tasks, even
though it does not reach an optimal strategy.

\begin{table}
    \caption{Win rates over 1000 games for out-of-training-domain maps, for all methods. The map on which this method was trained on is indicated on the left. The best result is in bold, the best result out of the reinforcement learning methods is in italics.}
    \label{tbl:test_out_of_domain}
    \vspace*{-0.4cm}
    \begin{center}
      \begin{tabular}{llcccc}
        \toprule
        train map & test map & best heuristic & DQN & PG & ZO \\
        \midrule
        m15v16 & m5v5 & \textbf{.96} (wc/c) & \textbf{\textit{.96}} & .79 & .80 \\
        & m15v15 & \textbf{.97} (c) & .27 & .16 & \textit{.80} \\
        & m18v18 & \textbf{.98} (c/noop) & .18 & .25 & \textit{.82} \\
        & m18v20 & \textbf{.63} (noop) & .00 & .01 & \textit{.17} \\
        \midrule
        w15v17 & w5v5 & \textbf{.78} (c) & .70 & .70 & \textit{.74} \\
        & w15v13 & 1. (rand\_nc/c) & 1. & .99 & 1. \\
        & w15v15 & .95 (c) & .87 & .61 & \textbf{\textit{.99}} \\
        & w18v18 & \textbf{.99} (c) & .92 & .56 & \textbf{\textit{1.}} \\
        & w18v20 & .71 (c) & .31 & .24 & .\textbf{\textit{76}} \\
        \bottomrule
      \end{tabular}
    \end{center}
\end{table}

\subsection{Interpretation of the learned policies}

We visually inspected the model's performance on large battles. On the larger
Marines map (m15v16), DQN learned to focus fire. Because this map has many
units, focus firing leads to units bumping into each other to try to focus on a
single unit. The PG player seemed to have a policy that attacks the closest
marine, though it doesn't do a good job switching targets. The Marines that are
not in range often bump into each other. Our zero order optimization learns a
hybrid between focus firing and attacking the closest unit. Units would switch
to other units in range if possible, but still focus on specific targets. This
leads to most Marines attacking constantly, as well as focus firing when they
can. However, the learned strategy was not perfected, since Marines would still
split their fire occasionally when left with few units.

In the Wraiths map (w15v17), the DQN player's strategy was hard to decipher.
The most likely explanation is that they tried to attack the closest target,
though it is likely the algorithm did not converge to a specific strategy. The
PG player learned to focus fire. However, because it only takes 6 Wraiths to
kill another, 9 actions are "wasted" (at the beginning
of the fight, when all our units are alive). Our zero order player learns that
focusing only on one enemy is not good, but it does not learn how many attacks
are necessary. This leads to a much higher win rate, but the player still
assigns more than 6 Wraiths to an enemy target (maybe for robustness to the
loss of one of our units), and occasionally will not focus fire when only a few
Wraiths are remaining.  This is similar to what the zero order player learned
during the Marines scenario.

\section{Conclusion}

This paper presents two main contributions. First, it establishes StarCraft
micromanagement scenarios as complex benchmarks for reinforcement learning:
with durative actions, delayed rewards, and large action spaces making random
exploration infeasible. Second, it introduces a new reinforcement learning
algorithm that performs better than prior work (DQN, PG) for discrete action
spaces in these micromanagement scenarios, with robust training (see
Figure~\ref{fig:training_uncertain}) and episodically consistent exploration
(exploring in the policy space).

This work leaves several doors open and calls for future work. Simpler
embedding models of state and actions, and variants of the model
presented here, have been tried, none of which produced efficient
units movement (e.g.  taking a unit out of the fight when its hit
points are low). There is ongoing work on convolutional networks based
models that conserve the 2D geometry of the game (while embedding the
discrete components of the state and actions). The zero order
optimization technique presented here should be studied more in depth,
and empirically evaluated on domains other than StarCraft
(e.g. Atari). As for StarCraft scenarios specifically, the subsequent
experiments will include self-play (training and evaluation),
multi-map training (training more generic models), and more complex
scenarios which include several types of advanced units with actions
other than move and attack. Finally, the goal of playing full games of
StarCraft should not get lost, so future scenarios would also include
the actions of ``recruiting'' units (deciding which types of unit to use),
as well as make use of them.

\section*{Acknowledgements}
We thank Y-Lan Boureau, Antoine Bordes, Florent Perronnin, Dave
Churchill, Léon Bottou and Alexander Miller for helpful discussions
and feedback about this work and earlier versions of the paper. We
thank Timoth\'{e}e Lacroix and Alex Auvolat for technical
contributions to our StarCraft/Torch bridge. We thank Davide Cavalca
for his support on Windows virtual machines in our cluster
environment.

\bibliographystyle{acm} 
\renewcommand{\baselinestretch}{0.9}
{\small
\bibliography{main}
}

\pagebreak

\renewcommand{\baselinestretch}{1}

\section{Appendix}

\subsection{StarCraft specifics}
\label{sec:specifics}

We advocate that using existing video games for RL experiments is interesting
because the simulators are oftentimes complex, and we (the AI programmers) do
not have control about the source code of the simulator. In RTS games like
StarCraft, we do not have access to a simulator (and writing one would be a
daunting task), so we cannot use (Monte Carlo) tree search
\cite{gelly2006exploration} directly, even less so in the setting of full games
\cite{ontanon2013survey}. In this paper, we consider the problem of
micromanagement scenarios, a subset of full RTS play.  Micromanagement is
about making good use of a given set of units in an RTS game. Units have
different features, like range, cooldown, hit points (health), attack power,
move speed, collision box etc.  These numerous features and the dynamics of the
game advantage player that take the right actions at the right times.
Specifically for the game(s) StarCraft, for which there are professional
players, very good competitive players and professional players perform more
than 300 actions per minute during intense battles.

We ran all our experiments on simple scenarios of battles of an
RTS game: StarCraft: Broodwar. These scenarios can be considered small scale
for StarCraft, but they already deem challenging for existing RL approaches.
For an example scenario of 15 units (that we control) against 16 enemy units,
even while reducing the action space to "atomic" actions (surrounding moves,
and attacks), we obtain 24 (8+16) possible discrete actions per unit for our
controller to choose from ($24^{15}$ actions total) at the beginning of the
battle. Battles last for tens of seconds, with durative actions, simultaneous
moves, and at 24 frames per second. The strategies that we need to learn
consist in coordinated sets of actions that may need to be repeated, e.g. focus
firing without overkill. We use a featurization that gives access only to the
state from the game, we do not pre-process the state to make it easier to learn
a given strategy, thus keeping the problem elegant and unbiased.

For most of these tasks (``maps''), the number of units that our RL agent has
to consider changes over an episode (a battle), as do its number of actions.
The fact that we are playing in this specific adversarial environment is that
if the units do not follow a coherent strategy for a sufficient amount of time,
they will suffer an unrecoverable loss, and the game will be in a state of the
game where the units will die very rapidly and make little damage,
independently of how they play -- a state that is mostly useless for learning.

Our tasks (``maps'') represent battles with homogeneous types of units, or with
little diversity (2 types of unit for each of the players).  For instance, they
may use a unit of type Marine, that is one soldier with 40 hit points, an
average move speed, an average range (approximately 10 times its collision
size), 15 frames of cooldown, 6 of attack power of normal damage type (so a
damage per second of 9.6 hit points per second, on a unit without armor).

On symmetric and/or monotyped maps, strategies that are required to win (on
average) are ``focus firing'', without overkill (not more units targeting a
unit than what is needed to kill it). For perfect win rates, some maps may
require that the AI moves its units out from the focus firing of the opponent.

\subsection{Hyper-parameters}
\label{sec:hyperparams}

Taking an action on every frame (24 times per second at the speed at
which human play StarCraft) for every unit would spam the game
needlessly, and it would actually prevent the units from
moving\footnote{Because several actions are durative, including
moves. Moves have a dynamic consisting of per-unit-type turn rate,
max speed, and acceleration parameters.}. We take actions for all
units synchronously on the same frame, even \texttt{skip\_frames}
frames. We tried several values of this hyper-parameter (5, 7, 9, 11,
13, 17) and we only saw smooth changes in performance. We ran all the
following experiments with a \texttt{skip\_frames} of 9 (meaning that
we take about 2.6 actions per unit per second). We also report the
strongest numbers for the baselines over all these skip frames. We
optimize all the models after each battle (episode), with RMSProp
(momentum $0.99$ or $0.95$), except for zero-order for which we
optimized with Adagrad (Adagrad did not seem to work better for DQN
nor REINFORCE). In any case, the learning rate was chosen among
$\{10^{-2}, 10^{-3}, 10^{-4}\}$.

For all methods, we tried experience replay, either with episodes (battles) as
batches (of sizes 20, 50, 100), or additionally with random batches of $(s_t,
a_t, r_{t+1}, s_{t+1}, terminal?)$ quintuplets in the case of Q-learning, it
did not seem to help compared to batching with the last battle. So, for
consistency, we only present results where the training batches consisted of
the last episode (battle).

For Q-learning (DQN), we tried two schemes of annealing for epsilon
greedy, $\epsilon = \frac{\epsilon_0}{\sqrt{1 + \epsilon_a
    . \epsilon_0 . t}}$ with $t$ the optimization batch, and $\epsilon
= \max(0.01, \frac{\epsilon_0}{\epsilon_a . t})$, Both with
$\epsilon_0 \in \{0.1, 1\}$, and respectively $\epsilon_a \in \{0,
\epsilon_0\}$ and $\epsilon_a \in \{10^{-5}, 10^{-4}, 10^{-3}\}$. We
found that the first works marginally better and used that in the
subsequent experiments with $\epsilon_0=1$ and $\epsilon_a=1$ for most
of the scenarios. We also used Double DQN as in \cite{van2015deep}
(thus implemented as target DQN). For the target/double network, we
used a lag of 100 optimizations, thus a lag of 100 battles in all the
following experiments. According to our initial runs/sweep, it seems
to slightly help for some cases of over-estimation of the Q value.

For REINFORCE we searched over $\tau \in \{0.1, 0.5, 1, 10\}$.

For zero-order, we tried $\delta\in\{0.1, 0.01, 0.001\}$.


\end{document}

%% file: macros.tex
\renewcommand{\card}{card\xfunction}

\newcommand{\act}{a}
\newcommand{\un}{u\xsubscript}
\newcommand{\cm}{c\xsubscript}
\newcommand{\ssize}[1]{|#1|}

\newcommand{\pol}{\pi\xfunction}
\newcommand{\parpol}{\pi_{\Theta}\xfunction}
\newcommand{\parpolpars}{\Theta}
\newcommand{\trsit}{\rho\xfunction}

\newcommand{\Units}{{\cal U}\xfunction}
\newcommand{\Cms}{{\cal C}}
\newcommand{\Acts}{{\cal A}\xfunction}

\newcommand{\lstun}[1]{\un{\ssize{#1}}}
\newcommand{\lstcm}[1]{\cm{\ssize{#1}}}

\newcommand{\rew}{r\xfunction}
\newcommand{\cumr}{\bar{r}\xfunction}
\newcommand{\Erew}{R\xfunction}
\newcommand{\origS}{{\cal S}}

\newcommand{\origs}{s}

\newcommand{\greeS}{\tilde{{\cal S}}}
\newcommand{\grees}{\tilde{s}}
\newcommand{\grew}{\tilde{r}\xfunction}
\newcommand{\gActs}{{\cal A}\xfunction}
\newcommand{\gtrsit}{\tilde{\rho}\xfunction}

\newcommand{\qq}{Q\xfunction}

\newcommand{\nr}{\bar{n}}
\newcommand{\sca}{z\xfunction}

\newcommand{\sign}{sign}





%% file: arxiv_long_version.bbl
\begin{thebibliography}{10}

\bibitem{abel2016exploratory}
{\sc Abel, D., Agarwal, A., Diaz, F., Krishnamurthy, A., and Schapire, R.~E.}
\newblock Exploratory gradient boosting for reinforcement learning in complex
  domains.
\newblock {\em arXiv preprint arXiv:1603.04119\/} (2016).

\bibitem{barto1983neuronlike}
{\sc Barto, A.~G., Sutton, R.~S., and Anderson, C.~W.}
\newblock Neuronlike adaptive elements that can solve difficult learning
  control problems.
\newblock {\em IEEE transactions on systems, man, and cybernetics}, 5 (1983),
  834--846.

\bibitem{bernstein2002complexity}
{\sc Bernstein, D.~S., Givan, R., Immerman, N., and Zilberstein, S.}
\newblock The complexity of decentralized control of markov decision processes.
\newblock {\em Mathematics of operations research 27}, 4 (2002), 819--840.

\bibitem{busoniu2008comprehensive}
{\sc Busoniu, L., Babuska, R., and De~Schutter, B.}
\newblock A comprehensive survey of multiagent reinforcement learning.
\newblock {\em IEEE Transactions on Systems, Man, And Cybernetics-Part C:
  Applications and Reviews, 38 (2), 2008\/} (2008).

\bibitem{churchill2012fast}
{\sc Churchill, D., Saffidine, A., and Buro, M.}
\newblock Fast heuristic search for rts game combat scenarios.
\newblock In {\em AIIDE\/} (2012).

\bibitem{clevert2015fast}
{\sc Clevert, D.-A., Unterthiner, T., and Hochreiter, S.}
\newblock Fast and accurate deep network learning by exponential linear units
  (elus).
\newblock {\em arXiv preprint arXiv:1511.07289\/} (2015).

\bibitem{deisenroth2013survey}
{\sc Deisenroth, M.~P., Neumann, G., and Peters, J.}
\newblock A survey on policy search for robotics.
\newblock {\em Foundations and Trends in Robotics 2}, 1-2 (2013), 1--142.

\bibitem{duchi2011adaptive}
{\sc Duchi, J., Hazan, E., and Singer, Y.}
\newblock Adaptive subgradient methods for online learning and stochastic
  optimization.
\newblock {\em Journal of Machine Learning Research 12}, Jul (2011),
  2121--2159.

\bibitem{duchi2013optimal}
{\sc Duchi, J.~C., Jordan, M.~I., Wainwright, M.~J., and Wibisono, A.}
\newblock Optimal rates for zero-order convex optimization: the power of two
  function evaluations.
\newblock {\em arXiv preprint arXiv:1312.2139\/} (2013).

\bibitem{gelly2006exploration}
{\sc Gelly, S., and Wang, Y.}
\newblock Exploration exploitation in go: Uct for monte-carlo go.
\newblock In {\em NIPS: Neural Information Processing Systems Conference
  On-line trading of Exploration and Exploitation Workshop\/} (2006).

\bibitem{ghadimi2013stochastic}
{\sc Ghadimi, S., and Lan, G.}
\newblock Stochastic first-and zeroth-order methods for nonconvex stochastic
  programming.
\newblock {\em SIAM Journal on Optimization 23}, 4 (2013), 2341--2368.

\bibitem{ghavamzadeh2006hierarchical}
{\sc Ghavamzadeh, M., Mahadevan, S., and Makar, R.}
\newblock Hierarchical multi-agent reinforcement learning.
\newblock {\em Autonomous Agents and Multi-Agent Systems 13}, 2 (2006),
  197--229.

\bibitem{hu1998multiagent}
{\sc Hu, J., and Wellman, M.~P.}
\newblock Multiagent reinforcement learning: theoretical framework and an
  algorithm.
\newblock In {\em ICML\/} (1998), vol.~98, pp.~242--250.

\bibitem{kiefer1952stochastic}
{\sc Kiefer, J., Wolfowitz, J., et~al.}
\newblock Stochastic estimation of the maximum of a regression function.
\newblock {\em The Annals of Mathematical Statistics 23}, 3 (1952), 462--466.

\bibitem{littman1994markov}
{\sc Littman, M.~L.}
\newblock Markov games as a framework for multi-agent reinforcement learning.
\newblock In {\em Proceedings of the eleventh international conference on
  machine learning\/} (1994), vol.~157, pp.~157--163.

\bibitem{liu2014evolving}
{\sc Liu, S., Louis, S.~J., and Ballinger, C.}
\newblock Evolving effective micro behaviors in rts game.
\newblock In {\em Computational Intelligence and Games (CIG), 2014 IEEE
  Conference on\/} (2014), IEEE, pp.~1--8.

\bibitem{maes2009structured}
{\sc Maes, F., Denoyer, L., and Gallinari, P.}
\newblock Structured prediction with reinforcement learning.
\newblock {\em Machine learning 77}, 2-3 (2009), 271--301.

\bibitem{mannor2003cross}
{\sc Mannor, S., Rubinstein, R.~Y., and Gat, Y.}
\newblock The cross entropy method for fast policy search.
\newblock In {\em ICML\/} (2003), pp.~512--519.

\bibitem{mnih2013playing}
{\sc Mnih, V., Kavukcuoglu, K., Silver, D., Graves, A., Antonoglou, I.,
  Wierstra, D., and Riedmiller, M.}
\newblock Playing atari with deep reinforcement learning.
\newblock In {\em Proceedings of NIPS\/} (2013).

\bibitem{mnih2015human}
{\sc Mnih, V., Kavukcuoglu, K., Silver, D., Rusu, A.~A., Veness, J., Bellemare,
  M.~G., Graves, A., Riedmiller, M., Fidjeland, A.~K., Ostrovski, G., et~al.}
\newblock Human-level control through deep reinforcement learning.
\newblock {\em Nature 518}, 7540 (2015), 529--533.

\bibitem{nemirovsky1982problem}
{\sc Nemirovsky, A.-S., Yudin, D.-B., and Dawson, E.-R.}
\newblock Problem complexity and method efficiency in optimization.

\bibitem{oh2016control}
{\sc Oh, J., Chockalingam, V., Singh, S., and Lee, H.}
\newblock Control of memory, active perception, and action in minecraft.
\newblock {\em arXiv preprint arXiv:1605.09128\/} (2016).

\bibitem{ontanon2013survey}
{\sc Ontan{\'o}n, S., Synnaeve, G., Uriarte, A., Richoux, F., Churchill, D.,
  and Preuss, M.}
\newblock A survey of real-time strategy game ai research and competition in
  starcraft.
\newblock {\em Computational Intelligence and AI in Games, IEEE Transactions on
  5}, 4 (2013), 293--311.

\bibitem{osband2016deep}
{\sc Osband, I., Blundell, C., Pritzel, A., and Van~Roy, B.}
\newblock Deep exploration via bootstrapped dqn.
\newblock {\em arXiv preprint arXiv:1602.04621\/} (2016).

\bibitem{osband2016generalization}
{\sc Osband, I., Roy, B.~V., and Wen, Z.}
\newblock Generalization and exploration via randomized value functions.
\newblock In {\em Proceedings of The 33rd International Conference on Machine
  Learning\/} (2016), pp.~2377--2386.

\bibitem{ruckstiess2010exploring}
{\sc R{\"u}ckstiess, T., Sehnke, F., Schaul, T., Wierstra, D., Sun, Y., and
  Schmidhuber, J.}
\newblock Exploring parameter space in reinforcement learning.
\newblock {\em Paladyn, Journal of Behavioral Robotics 1}, 1 (2010), 14--24.

\bibitem{sehnke2008policy}
{\sc Sehnke, F., Osendorfer, C., R{\"u}ckstie{\ss}, T., Graves, A., Peters, J.,
  and Schmidhuber, J.}
\newblock Policy gradients with parameter-based exploration for control.
\newblock In {\em Artificial Neural Networks-ICANN 2008}. Springer, 2008,
  pp.~387--396.

\bibitem{silver2016mastering}
{\sc Silver, D., Huang, A., Maddison, C.~J., Guez, A., Sifre, L., Van
  Den~Driessche, G., Schrittwieser, J., Antonoglou, I., Panneershelvam, V.,
  Lanctot, M., et~al.}
\newblock Mastering the game of go with deep neural networks and tree search.
\newblock {\em Nature 529}, 7587 (2016), 484--489.

\bibitem{silver2014deterministic}
{\sc Silver, D., Lever, G., Heess, N., Degris, T., Wierstra, D., and
  Riedmiller, M.}
\newblock Deterministic policy gradient algorithms.
\newblock In {\em ICML\/} (2014).

\bibitem{spall1997one}
{\sc Spall, J.~C.}
\newblock A one-measurement form of simultaneous perturbation stochastic
  approximation.
\newblock {\em Automatica 33}, 1 (1997), 109--112.

\bibitem{stone1999team}
{\sc Stone, P., and Veloso, M.}
\newblock Team-partitioned, opaque-transition reinforcement learning.
\newblock In {\em Proceedings of the third annual conference on Autonomous
  Agents\/} (1999), ACM, pp.~206--212.

\bibitem{sukhbaatar2016learning}
{\sc Sukhbaatar, S., Szlam, A., and Fergus, R.}
\newblock Learning multiagent communication with backpropagation.
\newblock {\em arXiv preprint arXiv:1605.07736\/} (2016).

\bibitem{sutton1988learning}
{\sc Sutton, R.~S.}
\newblock Learning to predict by the methods of temporal differences.
\newblock {\em Machine learning 3}, 1 (1988), 9--44.

\bibitem{sutton1998reinforcement}
{\sc Sutton, R.~S., and Barto, A.~G.}
\newblock {\em Reinforcement learning: An introduction}.
\newblock MIT press, 1998.

\bibitem{sutton1999policy}
{\sc Sutton, R.~S., McAllester, D.~A., Singh, S.~P., Mansour, Y., et~al.}
\newblock Policy gradient methods for reinforcement learning with function
  approximation.
\newblock In {\em NIPS\/} (1999), vol.~99, pp.~1057--1063.

\bibitem{synnaeve2011bayesian}
{\sc Synnaeve, G., and Bessiere, P.}
\newblock A bayesian model for rts units control applied to starcraft.
\newblock In {\em Computational Intelligence and Games (CIG), 2011 IEEE
  Conference on\/} (2011), IEEE, pp.~190--196.

\bibitem{szita2006learning}
{\sc Szita, I., and L{\"o}rincz, A.}
\newblock Learning tetris using the noisy cross-entropy method.
\newblock {\em Neural computation 18}, 12 (2006), 2936--2941.

\bibitem{tan1993multi}
{\sc Tan, M.}
\newblock Multi-agent reinforcement learning: Independent vs. cooperative
  agents.
\newblock In {\em Proceedings of the tenth international conference on machine
  learning\/} (1993), pp.~330--337.

\bibitem{tesauro1995temporal}
{\sc Tesauro, G.}
\newblock Temporal difference learning and td-gammon.
\newblock {\em Communications of the ACM 38}, 3 (1995), 58--68.

\bibitem{tesauro2003extending}
{\sc Tesauro, G.}
\newblock Extending q-learning to general adaptive multi-agent systems.
\newblock In {\em Advances in neural information processing systems\/} (2003),
  p.~None.

\bibitem{thompson1933likelihood}
{\sc Thompson, W.~R.}
\newblock On the likelihood that one unknown probability exceeds another in
  view of the evidence of two samples.
\newblock {\em Biometrika 25}, 3/4 (1933), 285--294.

\bibitem{Tieleman2012}
{\sc Tieleman, T., and Hinton, G.}
\newblock {Lecture 6.5---RmsProp: Divide the gradient by a running average of
  its recent magnitude}.
\newblock COURSERA: Neural Networks for Machine Learning, 2012.

\bibitem{van2015deep}
{\sc Van~Hasselt, H., Guez, A., and Silver, D.}
\newblock Deep reinforcement learning with double q-learning.
\newblock {\em arXiv preprint arXiv:1509.06461\/} (2015).

\bibitem{watkins1992q}
{\sc Watkins, C.~J., and Dayan, P.}
\newblock Q-learning.
\newblock {\em Machine learning 8}, 3-4 (1992), 279--292.

\bibitem{wender2012applying}
{\sc Wender, S., and Watson, I.}
\newblock Applying reinforcement learning to small scale combat in the
  real-time strategy game starcraft: broodwar.
\newblock In {\em Computational Intelligence and Games (CIG), 2012 IEEE
  Conference on\/} (2012), IEEE, pp.~402--408.

\bibitem{williams1992simple}
{\sc Williams, R.~J.}
\newblock Simple statistical gradient-following algorithms for connectionist
  reinforcement learning.
\newblock {\em Machine learning 8}, 3-4 (1992), 229--256.

\end{thebibliography}
